\documentclass[conference]{IEEEtran}
\IEEEoverridecommandlockouts
\usepackage{cite}
\usepackage{amsmath}
\usepackage{amssymb}
\usepackage{amsfonts}
\usepackage{algorithmic}
\usepackage{graphicx}
\usepackage{textcomp}
\usepackage{xcolor}
\usepackage{bm}
\usepackage{booktabs}
\usepackage{array}
\usepackage{caption}
\usepackage{dblfloatfix}
\usepackage{multirow}
\usepackage{booktabs}
\usepackage{pifont}
\usepackage{bbding}
\usepackage{hyperref}

\def\BibTeX{{\rm B\kern-.05em{\sc i\kern-.025em b}\kern-.08em
    T\kern-.1667em\lower.7ex\hbox{E}\kern-.125emX}}
\begin{document}

\title{SSAD: Self-supervised Auxiliary Detection Framework for Panoramic X-ray based Dental Disease Diagnosis\\
\thanks{Corresponding author: Linlin Shen and He Meng.}
}

\author{\IEEEauthorblockN{1\textsuperscript{st} Zijian Cai}
\IEEEauthorblockA{\textit{College of Computer Science and} \\
\textit{Software Engineering.}\\
\textit{Shenzhen University}\\
Shenzhen, China \\
zijianTsai@outlook.com}
\and
\IEEEauthorblockN{2\textsuperscript{nd} Xinquan Yang}
\IEEEauthorblockA{\textit{College of Computer Science and} \\
\textit{Software Engineering.}\\
\textit{Shenzhen University}\\
Shenzhen, China \\
xinquanyang99@gmail.com}
\and
\IEEEauthorblockN{3\textsuperscript{th} Xuguang Li}
\IEEEauthorblockA{\textit{Department of Stomatology} \\
\textit{Shenzhen University General Hospital}\\
Shenzhen, China \\
lixuguang@szu.edu.cn}
\and
\IEEEauthorblockN{4\textsuperscript{rd} Xiaoling Luo}
\IEEEauthorblockA{\textit{College of Computer Science and} \\
\textit{Software Engineering.}\\
\textit{Shenzhen University}\\
Shenzhen, China \\
xiaolingluoo@outlook.com}
\and
\IEEEauthorblockN{5\textsuperscript{rd} Xuechen Li}
\IEEEauthorblockA{\textit{School of Electronics and} \\
\textit{Information Engineering.}\\
\textit{Wuyi University}\\
Jiangmen, China \\
lixuechen1987622@126.com}
\and
\IEEEauthorblockN{6\textsuperscript{th} Linlin Shen}
\IEEEauthorblockA{\textit{College of Computer Science and} \\
\textit{Software Engineering.}\\
\textit{Shenzhen University}\\
Shenzhen, China \\
llshen@szu.edu.cn}
\and
\IEEEauthorblockN{7\textsuperscript{th} He Meng}
\IEEEauthorblockA{\textit{Department of Stomatology} \\
\textit{Shenzhen University General Hospital}\\
Shenzhen, China \\
menghe@szu.edu.cn}
\and
\IEEEauthorblockN{8\textsuperscript{th} Yongqiang Deng}
\IEEEauthorblockA{\textit{Department of Stomatology} \\
\textit{Shenzhen University General Hospital}\\
Shenzhen, China \\
qiangyongdeng@sina.com}
}


\maketitle

\begin{abstract}
Panoramic X-ray is a simple and effective tool for diagnosing dental diseases in clinical practice. When deep learning models are developed to assist dentist in interpreting panoramic X-rays, most of their performance suffers from the limited annotated data, which requires dentist's expertise and a lot of time cost. Although self-supervised learning (SSL) has been proposed to address this challenge, the two-stage process of pretraining and fine-tuning requires even more training time and computational resources. In this paper, we present a self-supervised auxiliary detection (SSAD) framework, which is plug-and-play and compatible with any detectors. It consists of a reconstruction branch and a detection branch. Both branches are trained simultaneously, sharing the same encoder, without the need for finetuning. The reconstruction branch learns to restore the tooth texture of healthy or diseased teeth, while the detection branch utilizes these learned features for diagnosis. To enhance the encoder's ability to capture fine-grained features, we incorporate the image encoder of SAM to construct a texture consistency (TC) loss, which extracts image embedding from the input and output of reconstruction branch, and then enforces both embedding into the same feature space. Extensive experiments on the public DENTEX dataset through three detection tasks demonstrate that the proposed SSAD framework achieves state-of-the-art performance compared to mainstream object detection methods and SSL methods. The code is available at \href{https://github.com/Dylonsword/SSAD}{https://github.com/Dylonsword/SSAD}. 
\end{abstract}

\begin{IEEEkeywords}
Dental Disease Detection, Deep Learning, Self Supervision
\end{IEEEkeywords}

\begin{figure}
\centering
\includegraphics[width=0.95\linewidth]{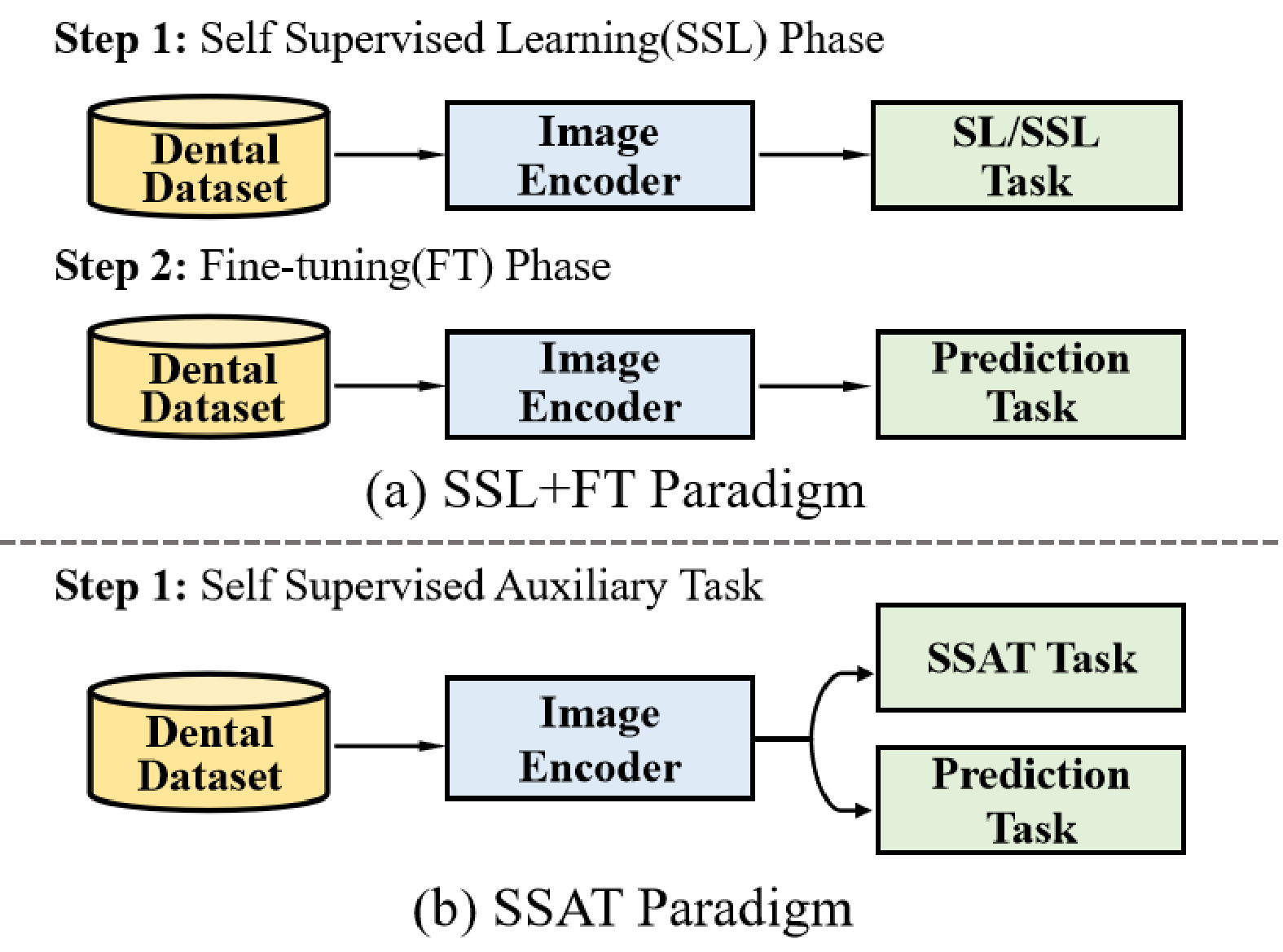}
\caption{Comparison of the SSL+FT paradigm and SSAT paradigm.} \label{SSL_FT_VS_SSAD}
\end{figure}

\section{Introduction}
Panoramic X-rays are widely used in dental diagnostics due to their fast imaging and cost-effectiveness. However, as the demand for precise treatment planning increases~\cite{yuksel2021dental}, dentists spend a significant amount of clinical time reading scans and visually interpreting panoramic X-rays~\cite{Kumar2021}. To address this challenge, introducing deep learning-based (DL) methods to automate the diagnostic process, is very useful. Clinically, a DL model should be able to automatically identify abnormal teeth and indicate the specific disease category, providing dentists with a valuable advantage in making quick decisions and saving their time.

\begin{figure*}
\centering
\includegraphics[width=0.95\linewidth]{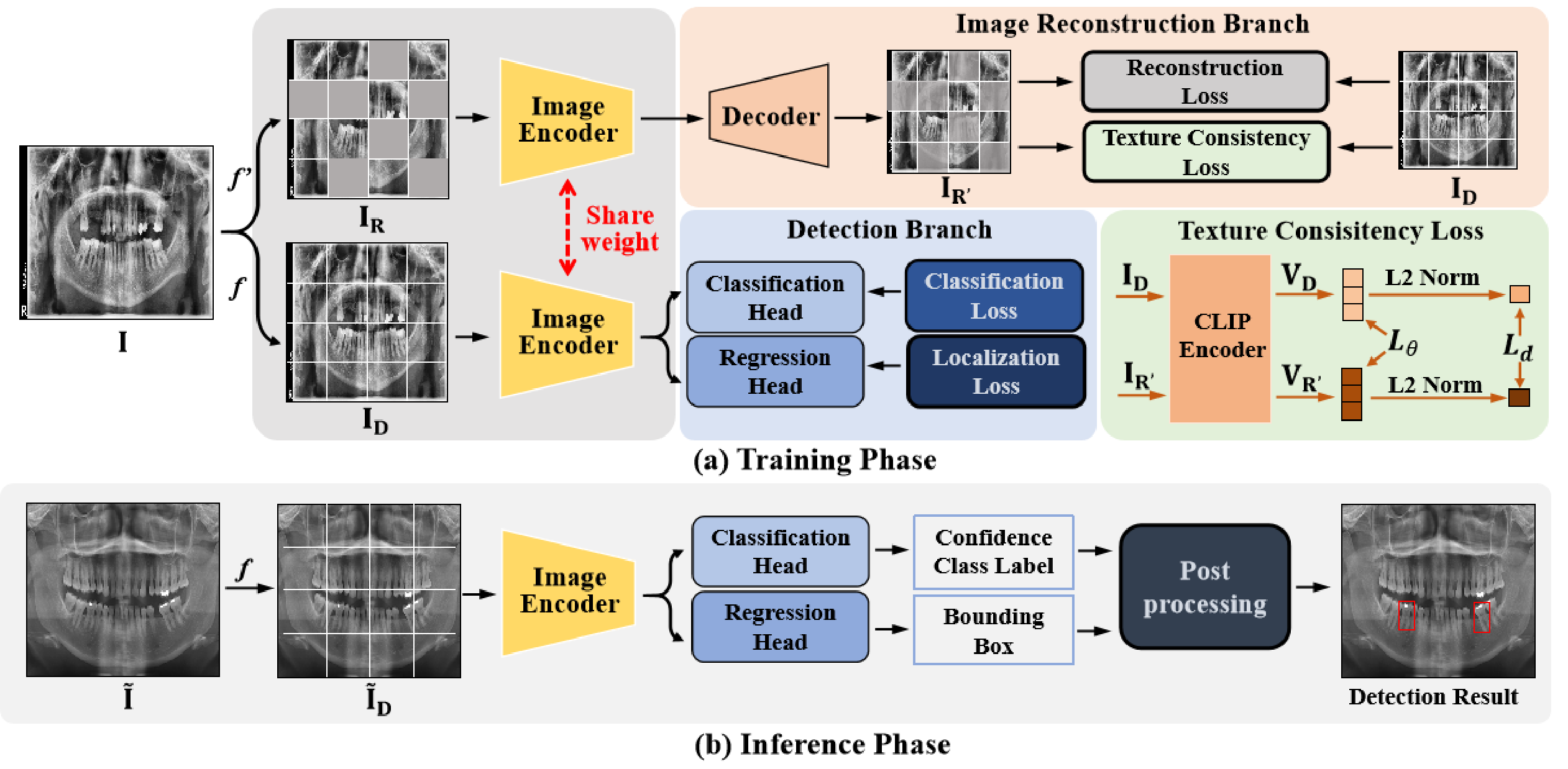}
\caption{The overview of the proposed self-supervised auxiliary detection framework.} \label{fig_network}
\end{figure*}

In recent years, deep learning (DL) models have been extensively employed to assist dentists in various tasks, including dental implant planning~\cite{yang2024simplify,yang2024implantformer,yang2023tceip,yang2023tcslot,yang2024two}, tooth detection~\cite{Chung2021}, dental enumeration~\cite{Lin2021,Tuzoff2019}, diagnosis of abnormalities~\cite{Zhu2022,Krois2019}, and treatment planning~\cite{yuksel2021dental}. Despite achieving some satisfactory results, these methods face challenge of limited availability of annotated data, which hinders further improvement in their performance. The annotation process for panoramic X-rays is complex and time-consuming, as it requires dentists to meticulously label each tooth with its corresponding disease category based on their expertise~\cite{Willemink2020}. Consequently, the challenge of improving network performance with limited annotation has become a widely discussed topic among researchers. 

Self-supervised learning (SSL) has proven to be an effective method to tackle this challenge~\cite{chen2020improved,grill2020bootstrap}, aiming to learn visual representations through pretext tasks, without annotation. Typically, the contrastive learning~\cite{he2020momentum} is employed to learn better image representation through the contrast between positive and negative image pairs. The learned weights are then transferred to downstream tasks through fine-tuning (FT). However, such two-stage process (SSL+FT) requires much more training time and computation-cost. Recently, researchers propose the self-supervised auxiliary task (SSAT)~\cite{das2024limited,li2022locality,liu2021efficient}, which jointly optimize the self-supervised task along with the primary task, e.g., detection or classification. By this means, the prediction network can be trained end-to-end without finetuning process (See Fig.~\ref{SSL_FT_VS_SSAD}), even achieves better performance than SSL methods.   

Draw inspiration from SSAT, in this paper, we develop a self-supervised auxiliary detection (SSAD) framework for diagnosing dental diseases. It consists of an image reconstruction branch and a disease detection branch. The image reconstruction branch aims to learn the visual representation of tooth texture, while the disease detection branch is designed to locate abnormal tooth and identify the corresponding disease category. Both branches share the same encoder to ensure consistent feature extraction. As the diagnosis of dental diseases relies on the tooth texture, we design a texture consistency (TC) loss to enable the encoder to capture more fine-grained features. Specifically, we introduce a strong feature extractor, e.g., the image encoder of SAM~\cite{radford2021learning} to extract image embedding from the input image and the restoration result of reconstruction branch. Then, both embedding are enforced to map into the same feature space. By this means, the detection branch can leverage the fine-grained feature from the encoder for better dental disease detection.


Main contributions of this paper can be summarized as follows: 1) To the best of our knowledge, the proposed SSAD is the first SSAT-based dental disease detection framework, which is plug-and-play and compatible with any detectors. (2) A texture consistency loss is designed to enable the encoder of SSAD to capture more fine-grained features, which greatly assists the detection branch to detect the dental diseases. (3) Extensive experiments on the public DENTEX dataset through three tasks demonstrated that the proposed SSAD achieves superior performance than available detection and SSL methods. 

\section{Method}
An overview of the proposed SSAD is presented in Fig.~\ref{fig_network}. It mainly consists of an image reconstruction branch and a disease detection branch. Given a patient's panoramic X-rays image $\mathbf{I}\in \mathbb{R}^{H\times W\times C}$, we first resize $\mathbf{I}$ to the fixed size and perform the random masking operation, to obtain masked image $\mathbf{I}_R\in \mathbb{R}^{H'\times W'\times C}$ and resized image $\mathbf{I}_D\in \mathbb{R}^{H'\times W'\times C}$. During training, the reconstruction branch and detection branch are trained simultaneously by sharing the same image encoder. The image encoder takes $\mathbf{I}_R$ and $\mathbf{I}_D$ as input and the output is fed into both branches for image reconstruction and disease detection. In inference, only the detection branch is retained for detecting the diseased tooth. Next, we will introduce both branches in detail.

\subsection{Reconstruction Branch}
In clinical, the diagnosis of dental disease relies on the tooth texture. Therefore, it's very important for an image encoder posses the capacity of capturing the fine-grained features to recognize the diseased or healthy teeth. To this end, we employ the widely adopted Simple Masked Image Modeling (simMIM) approach~\cite{xie2022simmim} as the reconstruction branch to restore the missing pixels from the masked panoramic X-rays image. By this means, the image encoder learns to restore the texture of healthy or diseased teeth, and these learned features is essential for the dental disease diagnosis. The architecture of reconstruction branch is shown in the Fig.\ref{fig_network}(a). Firstly, the input panoramic X-rays image $\mathbf{I}$ is resized to the fixed size ($512\times 512$ in this work) and the random masking operation $f'$ is performed (masking rate is set as 60\%). The masked image $\mathbf{I}_R$ is then fed into the image encoder for feature extraction. The decoder is comprised of several convolutional layers that upsample the encoder's output to produce the restoration result $\mathbf{I}_{R'}$. We follow simMIM to use the L1 loss to supervise the reconstruction process.

\subsection{Texture Consistency Loss}  As previously discussed, the reconstruction branch learns to restore the texture of healthy or diseased teeth for dental disease diagnosis. To further improve the restoration quality, we introduce the encoder of SAM to design an additional texture consistency (TC) loss to supervise the reconstruction branch. The overview of TC loss is given in Fig.\ref{fig_network}(a). It consists of a feature alignment loss $\mathcal{L}_{\theta}$ and a feature gather loss $\mathcal{L}_{d}$. Both losses aim to enable the reconstruction branch learns more robust feature to improve the restoration quality. Specifically, we introduce a strong feature extractor, i.e., the image encoder of CLIP, to extract the image embedding from $\mathbf{I}_{D}$ and $\mathbf{I}_{R'}$:
\begin{equation}
\begin{aligned}
    \bm{v}_{D} &= \textbf{E}(\mathbf{I}_{D}), \\
    \bm{v}_{R'} &= \textbf{E}(\mathbf{I}_{R'}),
\end{aligned}
\end{equation}
where $\textbf{E}$ represents the feature extractor. After obtaining the image embedding $\bm{v}_{D}$ and $\bm{v}_{R'}$, we first use $\mathcal{L}_{\theta}$ to align both vectors into the same feature angle: 
\begin{equation}
    \mathcal{L}_{\theta} = 1 - {D}_{cos}(\bm{v}_{D}, \bm{v}_{R'}), 
\end{equation}
where ${{D}_{cos}(\cdot)}$ represents cosine similarity. After performing the feature alignment, we use $\mathcal{L}_{d}$ to gather feature into the same feature space:
\begin{equation}
\mathcal{L}_{d} = \left| \| \bm{v}_{D} \|_2 - \| \bm{v}_{R'} \|_2 \right|,
\end{equation}
where $\bm{v}_{D}$ and $\bm{v}_{R'}$ are normalized by L2 normalization before conducting $\mathcal{L}_{d}$. The proposed TC loss is the combination of $\mathcal{L}_{\theta}$ and $\mathcal{L}_{d}$:
\begin{equation}
\mathcal{L}_{TC} = \mathcal{L}_{\theta} + \mathcal{L}_{d},
\end{equation}

\subsection{Detection Branch}
The detection branch is designed to locate abnormal tooth and identify the corresponding disease category. The architecture of the detection branch is shown in Fig.~\ref{fig_network}(a). It mainly consist of a image encoder, a classification head and a regression head. Firstly, the image encoder extracts features from $\mathbf{I}_{D}$ and the output will be fed into both prediction heads to locate the abnormal tooth. The regression head predicts the bounding box of diseased tooth and the classification head generates corresponding confidence scores and diseased category. 
The cross entropy loss and L1 loss are used to supervise the classification head and regression head, respectively.


\subsection{Self-supervised Auxiliary Detection Framework}
SSAD is proposed to improve the dental disease detection performance of detectors, which is plug-and-play and compatible with any detectors. It integrate a detection branch and an image reconstruction branch. During training, both branches train simultaneously by sharing the same image encoder. The reconstruction branch learns to restore the texture of healthy or diseased teeth from the masked image $\mathbf{I}_R$, which assists the detection branch better identify abnormal teeth. In the mean time, the encoder of detection branch extracts features from $\mathbf{I}_D$ to directly learn the image information. In inference, only the detection branch is retained and the standard detection process is performed. The architecture of SSAD is shown in  Fig.~\ref{fig_network}.


\begin{table}[h!]
\caption{The number of images for each task type.}\label{table1}
\centering
\begin{tabular}{c|c|c|c}
\toprule
\multicolumn{1}{c|}{Task} & \multicolumn{1}{c|}{Training} & \multicolumn{1}{c|}{Testing} & \multicolumn{1}{c}{Total} \\[1pt] \hline
\multicolumn{1}{c|}{Quadrant Detection} & \multicolumn{1}{c|}{554} & \multicolumn{1}{c|}{139} & \multicolumn{1}{c}{693} \\[1pt] \hline
\multicolumn{1}{c|}{Tooth Enumeration} & \multicolumn{1}{c|}{507} & \multicolumn{1}{c|}{127} & \multicolumn{1}{c}{634}\\[1pt] \hline
\multicolumn{1}{c|}{Disease Diagnosis} & \multicolumn{1}{c|}{564} & \multicolumn{1}{c|}{141} & \multicolumn{1}{c}{705} \\
\bottomrule
\end{tabular}
\end{table}

\section{Experiments and Results}
\subsection{Dataset and Implementation Details}
We use the publicly available dental disease dataset - DENTEX~\cite{hamamci2023diffusion} to evaluate the performance of SSAD. The dataset mainly consists of panoramic X-rays collected from patients above 12 years of age using the VistaPano S X-ray unit (Durr Dental, Germany). DENTEX contain three detection tasks: quadrant detection, tooth enumeration and disease diagnosis. For the diagnosis, four dental diseases is included, i.e., caries, deep caries, periapical lesions, and impacted teeth. We randomly select 80\% panoramic X-ray as train set and the remaining 20\% as test set. The data distribution of each task is given in Table~\ref{table1}.

Pytorch is used for SSAD framework training and testing. For the training of reconstruction branch, we use a batch size of 8, SGD optimizer and a learning rate with 2.5e-5 for DiffusionDet and HierachicalDet, and 1e-4 for other detectors. The encoder of CLIP is ViT-B/32, and the same data augmentation methods in~\cite{9879075} are applied. The training epoch of our SSAD framework is set as 100, and the SSL methods are set as 200 due to their slow convergence. For the detection branch's training, we employ the AdamW optimizer for each detector with a learning rate of 2.5e-5 for DiffusionDet and HierarchicalDet, and 1e-3 for other detectors. The learning rate is reduced by a factor of 10 at the 26th and 72nd epochs, respectively. All experiments are conducted on the NVIDIA A40 GPU platform. 

\begin{figure}
\centering
\includegraphics[width=1.0\linewidth]{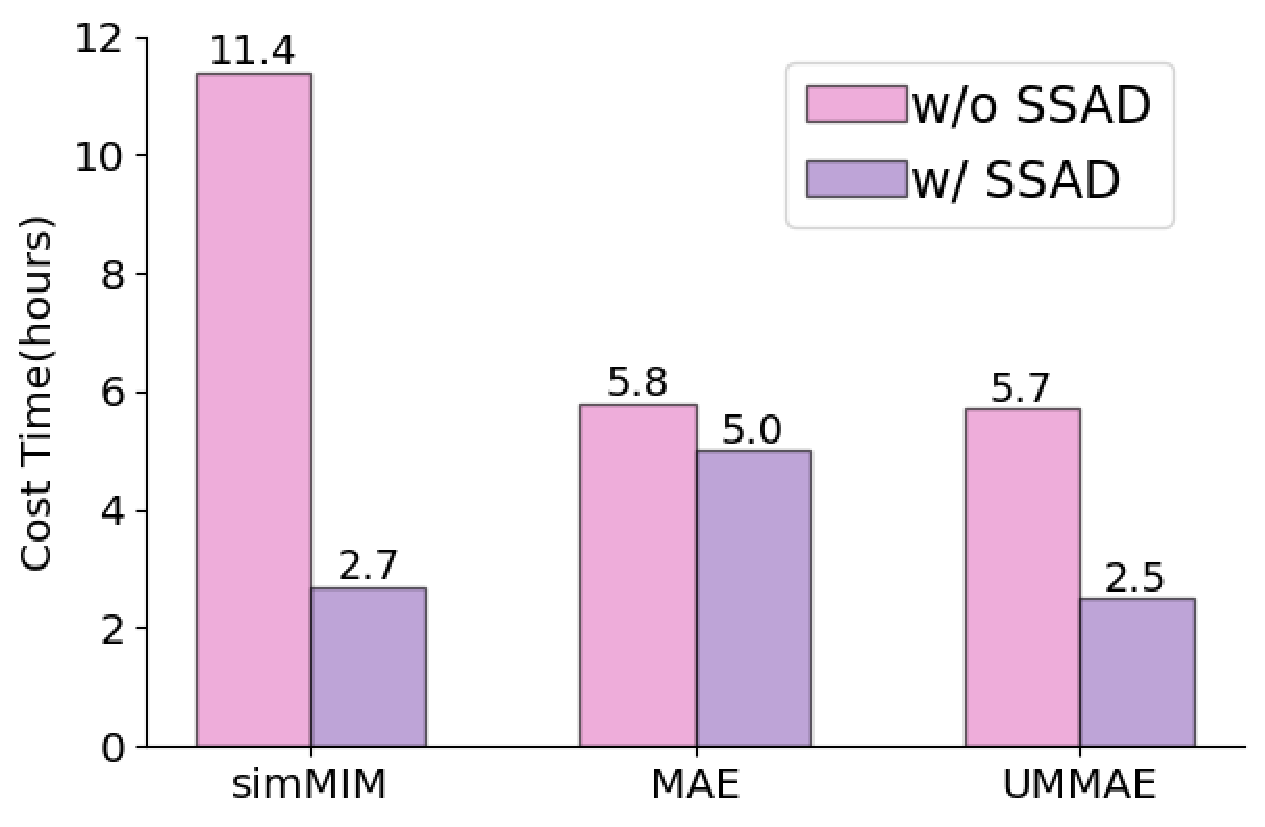}
\caption{The cost time of different paradigm on the task of disease diagnosis in training phase.} \label{ssad_vs_sslft_times}
\end{figure}

\subsection{Evaluation Criteria}
The general criterion of object detection, i.e., 
(mAP) is used to evaluate the predicting results. The $AP_{50}$, $AP_{75}$ and $AP_{50:95}$ denote the AP value of IoU=0.5, IoU=0.75 and the average IoU from 0.5 to 0.95, respectively. Moreover, more metrics are used to further validate the effectiveness of our method, i.e., AUC and Specificity. The calculation of these metrics are defined as follows:
\begin{equation}
Precition=\frac{TP}{TP+FP}
\end{equation}
\begin{equation}
Recall=\frac{TP}{TP+FN}
\end{equation}
\begin{equation}
mAP = \frac{1}{N} \sum_{i=1}^{N} \left( \int_0^1 P_i(r) \, dr \right)
\
\end{equation}
\begin{equation}
AUC=\frac{TP+TN}{TP+FN+FP+TN}
\end{equation}
\begin{equation}
Specificity=\frac{TN}{TN+FP}
\end{equation}

\begin{figure}
\centering
\includegraphics[width=1.0\linewidth]{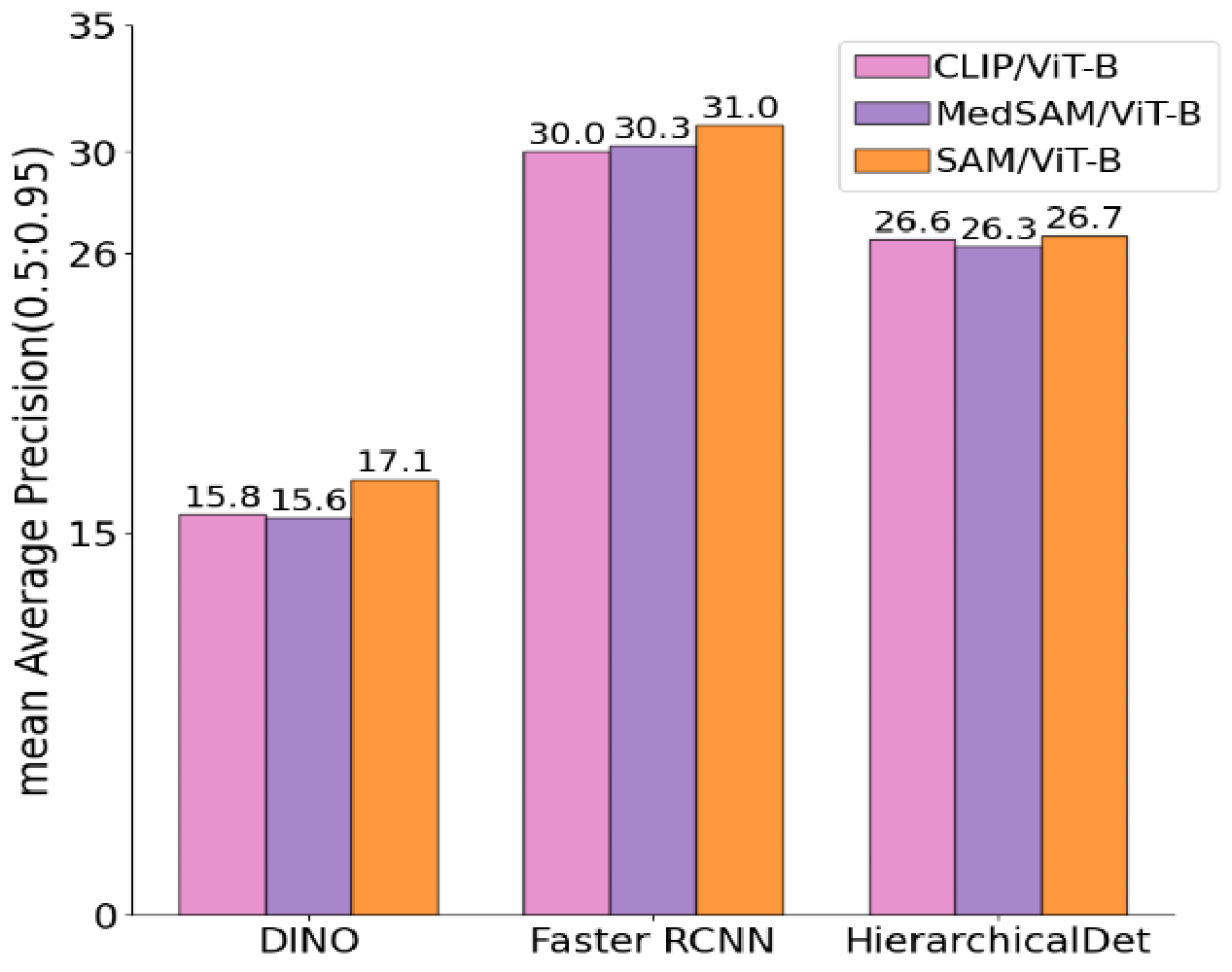}
\caption{The AP value of different detection methods using different feature extractors in TC loss.} \label{ssad_tc_backbone}
\end{figure}
Here, N is the number of disease categories. TP, FP and FN are the number of correct, false and missed predictions, respectively. P(r) is the PR Curve where the recall and precision act as abscissa and ordinate, respectively.

\begin{figure*}
\centering
\includegraphics[width=0.9\linewidth]{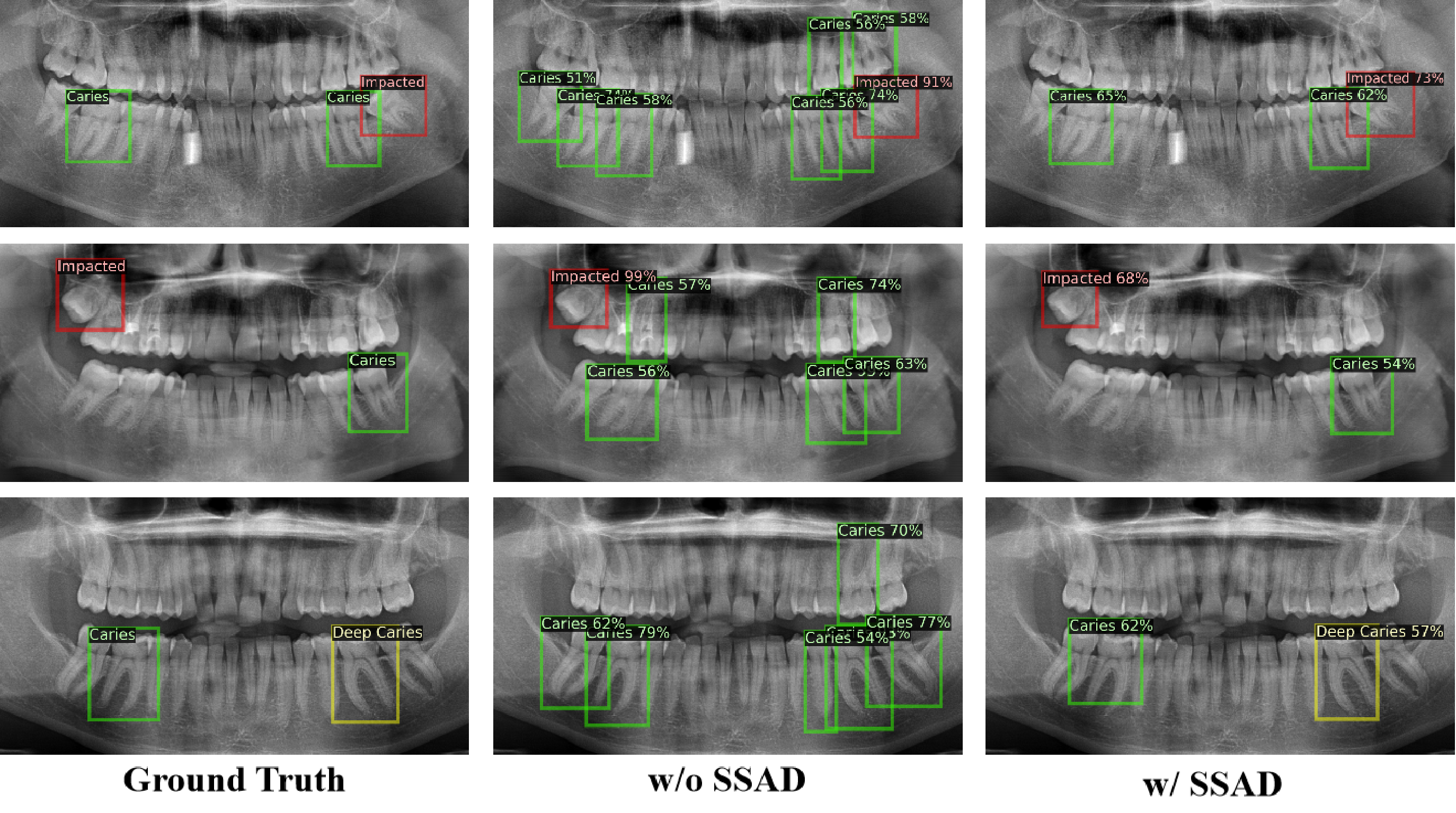}
\caption{Detection results of detector with or without SSAD.} \label{fig_vis}
\end{figure*}

\begin{table*}[t!] 
\caption{AP value (\%) of different paradigm on the task of disease diagnosis.}\label{table2}
\centering
\begin{tabular}{c|c|ccc}
\toprule
Paradigm & Method & \multicolumn{1}{c}{$AP_{50:95}$} & \multicolumn{1}{c}{$AP_{50}$} & \multicolumn{1}{c}{$AP_{75}$}\\ \hline
\multirow{3}{*}{SSL+FT}
 & simMIM & 12.93 & 24.95 & 10.07\\ 
 & MAE & 13.25 & 25.83 & 11.66\\
 & UMMAE & 13.31 & 25.67 & 12.93\\ \hline
\multirow{3}{*}{SSAD}
 & simMIM & $14.66(1.73 \uparrow)$ & $27.74(2.79 \uparrow)$ & $12.23(2.16 \uparrow)$\\ 
 & MAE & {${17.28(4.03 \uparrow)}$} & {${32.23(6.40 \uparrow)}$} & {${15.87(4.21 \uparrow)}$}\\ 
 & UMMAE & $16.52(3.21 \uparrow)$ & $31.87(6.20 \uparrow)$ & $13.45(0.52 \uparrow)$\\
\bottomrule
\end{tabular}
\end{table*}









\subsection{Performance Analysis}
\textbf{Comparison to the SSL Methods.} To demonstrate that the proposed SSAD performs better than the SSL methods, we conduct experiments to compare the AP value and training time between SSL and SSAD, results are given in Table~\ref{table2} and Fig~\ref{ssad_vs_sslft_times}. All experiments are conducted on the task of disease diagnosis. From Table~\ref{table2} we can observe that SSAD brings improvement for all SSL methods, in which MAE achieves the best performance - 17.28\% , while only 13.25\% for the SSL paradigm. In terms of training time, SSAD largely reduces the training time, i.e. the training time for SSAD and SSL is 2.7 hours and 11.4 hours, respectively. These results greatly demonstrate the effectiveness of SSAD.

\textbf{Comparison of Different Feature Extractors.} In Fig.~\ref{ssad_tc_backbone}, we compare the AP value of different detection methods when using different feature extractors in TC loss. We select three detectors (DINO, Faster RCNN and HierachicalDet) which uses different encoder in Table~\ref{table5}. From the figure we can observe that SAM achieves the best performance, while the performances of CLIP and MedSAM are similar in DINO and Faster RCNN. Therefore, we choose the SAM as the feature extractor in our TC loss.

\begin{table*}[t!]
\caption{AP value (\%) of different detection methods with or without SSAD.}\label{table3}
\centering
\begin{tabular}{c|c|ccccc|cccccc}
\toprule

\multicolumn{1}{c|}{\multirow{2}{*}{Network}} &
\multicolumn{1}{c|}{\multirow{2}{*}{Encoder}} &
\multicolumn{5}{c|}{w/o SSAD} & 
\multicolumn{6}{c}{w/ SSAD} \\
\multicolumn{1}{c|}{} & \multicolumn{1}{c|}{} & 
$AP_{50:95}$  & $AP_{50}$  & $AP_{75}$ & $AUC$ & \multicolumn{1}{c|}{$Spec$}  & 
$\mathcal{L}_{TC}$ &
$AP_{50:95}$    & $AP_{50}$    & $AP_{75}$ & $AUC$ & $Spec$   \\ \hline

\multicolumn{1}{c|}{\multirow{2}[2]{*}{DINO}} & 
\multicolumn{1}{c|}{\multirow{2}[2]{*}{ResNet-50}} 
& \multirow{2}[2]{*}{15.42} & \multirow{2}[2]{*}{28.26} & \multirow{2}[2]{*}{14.46} & \multirow{2}[2]{*}{89.49} & \multicolumn{1}{c|}{\multirow{2}[2]{*}{95.76}} 
& \ding{55}  & 15.57 & 27.96 & 15.50 & \textbf{92.53} & 96.35 \\
\multicolumn{1}{c|}{} &
\multicolumn{1}{c|}{} 
&  &  &  &  & \multicolumn{1}{c|}{} 
& \Checkmark & \textbf{17.12} & \textbf{30.03} & \textbf{16.42} & 89.35 & \textbf{96.38}  \\
\hline

\multicolumn{1}{c|}{\multirow{2}[2]{*}{Faster RCNN}} & 
\multicolumn{1}{c|}{\multirow{4}[3]{*}{ViTDet-B}} 
& \multirow{2}[2]{*}{28.93} & \multirow{2}[2]{*}{46.16} & \multirow{2}[2]{*}{33.68} & \multirow{2}[2]{*}{73.26} & \multicolumn{1}{c|}{\multirow{2}[2]{*}{81.36}} & \ding{55}
& 29.74 & 46.02 &  35.46 & 73.70 & 86.17 \\
\multicolumn{1}{c|}{} & 
\multicolumn{1}{c|}{} 
&  &  &  &  & \multicolumn{1}{c|}{} & \Checkmark
& \textbf{31.04} & \textbf{48.01} &  \textbf{36.83} & \textbf{79.33} & \textbf{86.34} \\
\cline{1-1} \cline{3-13}

\multicolumn{1}{c|}{\multirow{2}[2]{*}{FCOS}} & 
\multicolumn{1}{c|}{\multirow{2}[3]{*}{}} 
& \multirow{2}[2]{*}{14.46} & \multirow{2}[2]{*}{27.93} & \multirow{2}[2]{*}{12.18} & \multirow{2}[2]{*}{67.60} & \multicolumn{1}{c|}{\multirow{2}[2]{*}{\textbf{98.30}}} & \ding{55}
& 18.41 & 33.33 &  17.34 & 70.60 & 97.62 \\
\multicolumn{1}{c|}{} & 
\multicolumn{1}{c|}{} 
&  &  &  &  & \multicolumn{1}{c|}{} & \Checkmark
& \textbf{19.18} & \textbf{33.76} &  \textbf{21.47} & \textbf{71.35} & 97.26 \\
\hline

\multicolumn{1}{c|}{\multirow{2}[2]{*}{DiffusionDet}} & 
\multicolumn{1}{c|}{\multirow{4}[3]{*}{Swin-B}} 
& \multirow{2}[2]{*}{10.25} & \multirow{2}[2]{*}{23.04} & \multirow{2}[2]{*}{7.74} & \multirow{2}[2]{*}{\textbf{88.49}} & \multicolumn{1}{c|}{\multirow{2}[2]{*}{98.99}} & \ding{55}
& 10.30 & 24.21 &  7.20 & 79.55 & 99.90\\
\multicolumn{1}{c|}{} & 
\multicolumn{1}{c|}{} 
&  &  &  &  & \multicolumn{1}{c|}{} & \Checkmark
& \textbf{11.40} & \textbf{25.55} &  \textbf{8.34} & 79.57 & \textbf{99.97} \\
\cline{1-1} \cline{3-13}

\multicolumn{1}{c|}{\multirow{2}[2]{*}{HierarchicalDet}} & 
\multicolumn{1}{c|}{\multirow{2}[3]{*}{}} 
& \multirow{2}[2]{*}{24.15} & \multirow{2}[2]{*}{40.27} & \multirow{2}[2]{*}{27.72} & \multirow{2}[2]{*}{83.41} & \multicolumn{1}{c|}{\multirow{2}[2]{*}{99.16}} & \ding{55}
& 25.34 & 43.01 & 30.25 & \textbf{84.29} & 99.02 \\
\multicolumn{1}{c|}{} & 
\multicolumn{1}{c|}{} 
&  &  &  &  & \multicolumn{1}{c|}{} &  \Checkmark
& \textbf{26.70} & \textbf{43.97} & \textbf{30.36} & 83.67 & \textbf{99.30}  \\
\bottomrule
\end{tabular}%
\end{table*}

\begin{table*}[t!]
\caption{AP value (\%) of different detection tasks on the DENTEX dataset.}\label{table5}
\centering
\begin{tabular}{c|c|cccc}
\toprule

\multicolumn{1}{c|}{Network} &
\multicolumn{1}{c|}{Encoder} & SSAD
 &$AP_{50:95}$  & $AP_{50}$  & $AP_{75}$   \\ \hline

\multicolumn{6}{c}{Quadrant} \\ \hline
\multicolumn{1}{c|}{\multirow{2}[2]{*}{HierachicalDet}} & 
\multicolumn{1}{c|}{\multirow{2}[2]{*}{Swin-B}} & \ding{55}  & 68.63 & 98.87 & 85.51 \\
\multicolumn{1}{c|}{} & 
\multicolumn{1}{c|}{} & \Checkmark & \textbf{69.12} & \textbf{99.19} & \textbf{88.56} \\ \hline
\multicolumn{6}{c}{Enumeration} \\ \hline
\multicolumn{1}{c|}{\multirow{2}[2]{*}{HierachicalDet}} & 
\multicolumn{1}{c|}{\multirow{2}[2]{*}{Swin-B}} & \ding{55}  & 49.71 & 92.76 & 47.85 \\
\multicolumn{1}{c|}{} & 
\multicolumn{1}{c|}{} & \Checkmark & \textbf{50.27} & \textbf{92.88} & \textbf{48.08} \\

\bottomrule
\end{tabular}%
\end{table*}

\textbf{Ablation Study.} To further validate the effectiveness of the proposed SSAD framework, we compare the detection performance of different state-of-the-art detection methods in Table~\ref{table3}. Specifically, we use different backbone, e.g., CNN-based (ResNet-50~\cite{he2016deep}) and transformer-based (ViT~\cite{dosovitskiy2020image} and Swin Transformer~\cite{Liu_2021_ICCV}), and different network architecture, e.g., CNN-based (Faster RCNN~\cite{DBLP:journals/corr/RenHG015} and FCOS~\cite{DBLP:journals/corr/abs-1904-01355}), transformer-based (DINO~\cite{zhang2022dino}) and Diffusion-based (DiffusionDet~\cite{chen2022diffusiondet} and HierarchicalDet~\cite{hamamci2023diffusion}) for comparison. 

From the table we can observe that SSAD brings performance improvement for all detectors, in which FCOS achieves the largest improvement of AP value (3.95\%). When TC loss is introduced, these detector's performance will continue to increase. The CNN-based method, Faster RCNN achieves the best performance - 31.04\% among all detectors. This phenomenon demonstrates the effectiveness of SSAD, which is plug-and-play and compatible with any detectors. In terms of AUC and Specificity, the proposed SSAD also achieves superior performances.

In Fig.~\ref{fig_vis}, we visualize the detection results of detector with or without training by SSAD. We take the best performed Faster RCNN for visualization. The first column shows panoramic X-ray images with correct bounding boxes. The second and third columns display detection results without and with the SSAD framework, respectively. From the images, we can observe that detectors trained by supervise learning generates  numerous false positive detection, and they also make mistakes in distinguishing between similar categories (i.e., Caries and Deep Caries). When SSAD is introduced, the detection branch learns the texture feature of diseased teeth from the reconstruction branch, which assists the SSAD-based detector generates accurate detection results and are able to differentiate between similar categories more effectively.

\textbf{Different Detection Tasks.} To verify the generality of the proposed SSAD framework, we compare the AP value of detector trained with or without SSAD on different detection tasks (quadrant detection and tooth enumeration), results are given in Table~\ref{table5}. Specifically, we use the HierachicalDet as the detector, and Swin-B is set as the encoder. From the table we can observe that the proposed SSAD brings improvement for both detection tasks. This phenomenon indicates that the proposed SSAD can perform well on different detection tasks.

\section{Conclusions}
In this paper, we develop a self-supervised auxiliary detection (SSAD) framework, which consists of an image reconstruction branch and a disease detection branch. The reconstruction branch learns to restore the tooth texture to assist the detection branch diagnosing dental diseases. A texture consistency (TC) loss is designed to enable the image encoder to capture more fine-grained features. Extensive experiments on the public DENTEX dataset demonstrated that the proposed SSAD achieves superior performance than the existing methods.

\section*{Acknowledgment}
This work was supported by the National Natural Science Foundation of China under Grant 82261138629 and 12326610; Guangdong Basic and Applied Basic Research Foundation under Grant 2023A1515010688; Shenzhen Municipal Science and Technology Innovation Council under Grant JCYJ20220531101412030; Medicine-Engineering Interdisciplinary Research Foundation of ShenZhen University 00000351.

\bibliographystyle{splncs04}
\bibliography{ref}

\end{document}